\documentclass[letterpaper,10pt,conference]{IEEEtran}

\usepackage[T1]{fontenc}
\usepackage[english]{babel}
\usepackage{amsmath}
\usepackage{amsfonts}
\usepackage{amssymb}
\usepackage{bm}
\usepackage{xspace}
\usepackage[boxruled]{algorithm2e}
\usepackage{mathrsfs}
\usepackage{multirow}
\usepackage{tabularray}
\usepackage{eurosym}
\usepackage{dsfont}
\usepackage[normalem]{ulem} 
\usepackage{balance}
\usepackage{soul}
\usepackage{pdfpages}
\usepackage{graphicx}
\usepackage{todonotes}
\usepackage{url}
\usepackage{siunitx}
\usepackage{pgfplots}
\pgfplotsset{compat=1.18}
\usepackage{stfloats}
\usepackage{array}
\usepackage{adjustbox}
\usepackage{float}
\allowdisplaybreaks
\usepackage{cite}
\usepackage{multibib}
\usepackage{makecell}
\usepackage{xfp}

\usepackage[hidelinks]{hyperref}

\newcommand{\up}[1]{^{\text{#1}}}
\newcommand{\down}[1]{_{\text{#1}}}
\newcommand{\norm}[1]{\left\lVert#1\right\rVert}
\newcommand{\R}{\mathbb{R}}
\newcommand{\N}{\mathbb{N}}

\newcommand{\SE}{SE(3)}

\makeatletter
\let\old@ps@headings\ps@headings
\let\old@ps@IEEEtitlepagestyle\ps@IEEEtitlepagestyle
\def\confheader#1{%
    \def\ps@headings{%
        \old@ps@headings%
        \def\@oddhead{\strut\hfill#1\hfill\strut}%
        \def\@evenhead{\strut\hfill#1\hfill\strut}%
    }%
    \def\ps@IEEEtitlepagestyle{%
        \old@ps@IEEEtitlepagestyle%
        \def\@oddhead{\strut\hfill#1\hfill\strut}%
        \def\@evenhead{\strut\hfill#1\hfill\strut}%
    }%
    \ps@headings%
}
\makeatother
\confheader{%
    \parbox{20cm}{\textit{Accepted Preprint at 2024 IEEE-RAS 23rd International Conference on Humanoid Robots (Humanoids), October 2024}}
}

\definecolor{lightyellow}{rgb}{1.0,0.98,0.7}
\sethlcolor{lightyellow}

\title{
Flow Matching Imitation Learning\\for Multi-Support Manipulation
}
\author{
Quentin Rouxel, Andrea Ferrari, Serena Ivaldi, and Jean-Baptiste Mouret
\thanks{
The authors are with Inria, CNRS, Universit\'e de Lorraine, France. {\tt\footnotesize firstname.lastname@inria.fr}. A. Ferrari is also with Sapienza University of Rome, Italy.}%
\thanks{This research was supported by the CPER CyberEntreprises, the Creativ’Lab platform of Inria/LORIA, the EU Horizon project euROBIN (GA n.101070596), the France 2030 program through projects PEPR O2R AS3 and PI3 (ANR-22-EXOD-007, ANR-22-EXOD-004).}%
}

\IEEEoverridecommandlockouts
\begin{document}

\maketitle

\begin{abstract}
Humanoid robots could benefit from using their upper bodies for support contacts, enhancing their workspace, stability, and ability to perform contact-rich and pushing tasks. In this paper, we propose a unified approach that combines an optimization-based multi-contact whole-body controller with Flow Matching, a recently introduced method capable of generating multi-modal trajectory distributions for imitation learning. In simulation, we show that Flow Matching is more appropriate for robotics than Diffusion and traditional behavior cloning. On a real full-size humanoid robot (Talos), we demonstrate that our approach can learn a whole-body non-prehensile box-pushing task and that the robot can close dishwasher drawers by adding contacts with its free hand when needed for balance. We also introduce a shared autonomy mode for assisted teleoperation, providing automatic contact placement for tasks not covered in the demonstrations. Full experimental videos are available at: \url{https://hucebot.github.io/flow_multisupport_website/}
\end{abstract}

\section{Introduction}

In spite of the many advances in whole-body control, the tasks of most current humanoid robots are implicitly split into two parts: feet for locomotion and support, and hands for manipulation and other interactions with the world. This view overlooks all the possible uses of arms as additional support as well as non-prehensile manipulation like pushing with the side of the forearm, sliding and, more generally using the body of the robot as a potential contact surface. By contrast, humans routinely lean on a table to grasp a distant object, push on a wall while pulling a heavy door, exploit handrails to increase their stability, keep a door open with their shoulder, etc.

\begin{figure}[t]
    \centering
    \includegraphics[trim=0cm 0cm 0cm 0cm,clip,width=0.99\linewidth]{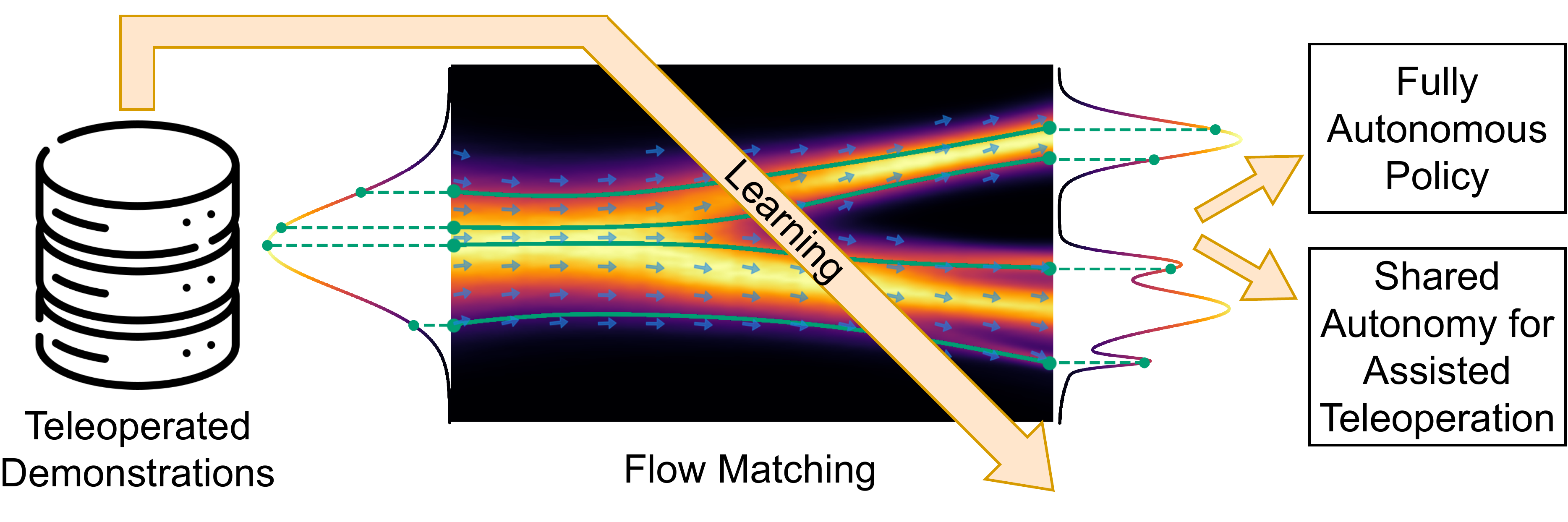}
	  \includegraphics[trim=0cm 0cm 0cm 0cm,clip,width=0.32\linewidth]{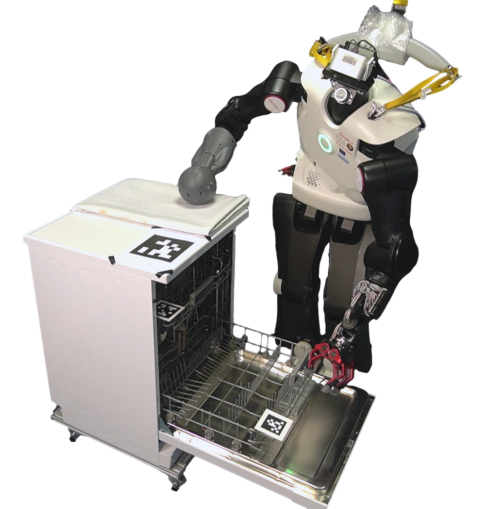}
    \includegraphics[trim=0cm 0cm 0cm 0cm,clip,width=0.32\linewidth]{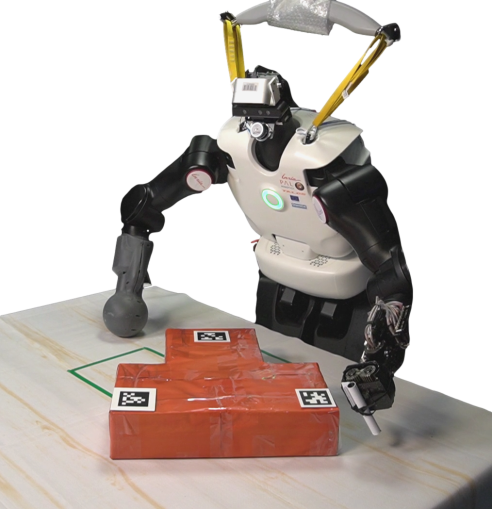}
 	\includegraphics[trim=0cm 0cm 0cm 0cm,clip,width=0.32\linewidth]{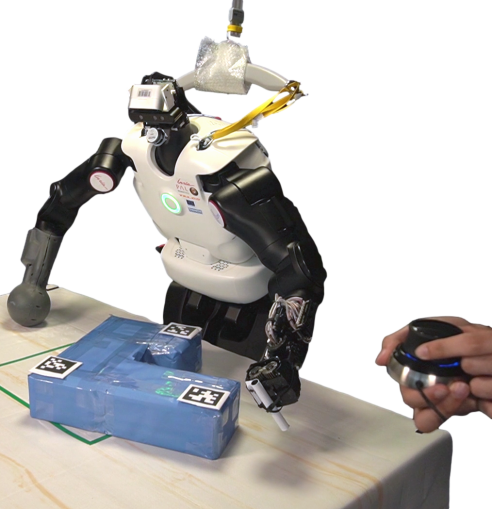}
    \caption{
    To perform multi-support tasks, the Talos humanoid robot uses its right hand as an additional support to extend its reach and maintain balance. Imitation learning allows the robot to autonomously solve these tasks or assist a human operator with automatic contact placement (see videos).
    }
    \label{fig:concept}
\end{figure}

In this work, we focus on these scenarios that leverage whole-body motion and multi-contact strategies to extend the manipulation capabilities (Fig.~\ref{fig:concept}). We term them \textit{multi-support manipulation tasks}, by analogy with the traditional single and double support cases for humanoids.

Our objective is to design control policies for humanoid robots that can leverage contacts when needed, both for adding support and perform non-prehensile tasks. On the one hand, model-based planners could search for support contacts, as this is often done with footstep planning \cite{tonneau2018efficient,ruscelli2020multi}, but this requires a very good understanding of the world, as many surfaces are not suitable contact surface (fragile surfaces like windows, slippery surfaces, ...). On the other hand, model-based approaches do not work well for pushing or sliding tasks because of the non-linear dynamics of sliding and friction \cite{woodruff2017planning,moura2022non}.

In this work, we address these two challenges with a single, unified method: imitation learning for whole-body multi-support motions. Hence, by demonstrating when and how to establish contacts, we can leverage the human ``common sense'' to choose contacts, avoid modeling explicitly the environment and sliding dynamics, and achieve real-time performance. While imitation learning has been applied to many tasks, it has not yet been investigated, to the best of our knowledge, for whole-body multi-contact and contact switching scenarios.

Many approaches have been proposed for imitation learning in robotics. The most traditional approach is behavior cloning, in which a neural network is learned with supervised learning to associate states to actions \cite{argall2009survey}. To exploit the structure of trajectories and control, a popular approach has been Dynamic Motion Primitives \cite{schaal2006dynamic} and various extensions like Probabilistic Motion Primitives \cite{paraschos2013probabilistic}. However, these methods tend to not scale well to high-dimensional inputs, like images, and large datasets. Also they are typically unable to model multi-modal distributions of demonstrations, whereas multi-modality is critical for many humanoid tasks. For example, if a humanoid can reach two contact locations, left or right (Fig.~\ref{fig:task_reach}), to add an extra support for balance, then averaging all demonstrations assuming a unimodal distribution will result in the policy averaging left and right positions and placing the contact in-between the two, causing the robot to fall.

The recent successes on generative processes for images and sound, like DALL-E, have inspired a new breed of behavior cloning algorithms \cite{janner2022planning, visuo_diffusion}. In essence, instead of generating an image conditioned by a text input, these algorithms generate trajectories conditioned by a state. 

The heart of these generative algorithms is a diffusion process \cite{ddpm} that learns the probability distribution of actions demonstrated by human operators and then sample new actions from this learned model. Diffusion methods were recently connected to optimal transport theory \cite{albergo2023stochastic, albergo2023building}, and linked with flow-based methods \cite{flow} within a unified framework, where Diffusion represents the stochastic counterpart and Flow Matching the deterministic counterpart. In this work, we hypothetize that the flow-based approaches, specifically Flow Matching, is best suited for robotics applications: it offers a simpler framework than the initial diffusion approach, that can yields deterministic outputs, and allows for faster inference without loss of quality compared to Diffusion.

In this paper, we show that a policy trained from demonstrations can effectively provide useful assistance for multi-support manipulation tasks, especially in the automatic placement of contacts. We are interested in both autonomous task execution as well as in assisted teleoperation/shared autonomy \cite{li2023classification}, where a human operator controls one robot end-effector (e.g., the left hand) while the robot autonomously controls its entire body and notably autonomously determines support contacts (e.g., with the right hand), regulates contact forces, to make sure that the task commanded by the human can be executed without the robot falling.

In summary, the contributions of our work are three-fold:
\begin{itemize}
    \item We introduce an imitation learning formulation and architecture that enables multi-support manipulation tasks.
    \item We showcase the Flow Matching generative method for generating whole-body movements on a full-size humanoid robot, demonstrating its advantages over Diffusion methods and its potential for robotic applications.
    \item We demonstrate that the autonomous policy learned from demonstrations can assist the human operator in a shared autonomy mode. This assistance performs automatic contact placement and is valuable in situations where the task varies from the demonstrated scenario, making the policy unable to solve the task alone.
\end{itemize}

\section{Related Work}\label{sec:related_works}

Classical model-based approaches address multi-contact tasks hierarchically. Simplified template models or heuristics determine contact placement and sequence \cite{tonneau2018efficient}. Then trajectory planning \cite{ruscelli2020multi} and control methods generate optimized whole-body motions, tracking them on the actual system while regulating interaction forces and maintaining balance. The control of complex robots, humanoids, and multi-limb systems \cite{polverini2020multi}, is well understood through model-based optimization approaches and these methods have been demonstrated on both torque-controlled \cite{cisneros2020inverse,henze2016passivity, abi2019torque} and position-controlled \cite{samadi2021humanoid,hiraoka2021online,murooka2022centroidal,rouxel2024multicontact} robots. However, optimizing contact placement and sequencing is highly challenging, involving both continuous and discrete decisions. Contact-rich tasks, such as non-prehensile \cite{woodruff2017planning,moura2022non} tasks, pose significant challenges due to sliding contacts, diverse valid strategies of sequences, and the requirement to consider the entire object geometry rather than predefined contact points.






Instead, we adopted an imitation learning approach \cite{Billard2016} which learns from human demonstrations, and specifically the Behavior Cloning (BC) method \cite{argall2009survey} building a policy that directly maps observations to actions in a supervised manner. Traditional BC such as DMP \cite{schaal2006dynamic} or ProMPs \cite{paraschos2013probabilistic} performs well on simple tasks within the demonstrated state-space distribution but suffers from accumulation of prediction errors, which can lead to state divergence and failure. To address this, our policy predicts \textit{trajectories} of actions, aligning with recent works \cite{aloha_zhao_learning_2023, visuo_diffusion, 3d_diffusion}, enhancing temporal coherence and mitigating error compounding.

Another limitation of BC is handling the variability in human demonstrations, idle actions, and different strategies used to solve the same tasks. These demonstrations form a multi-modal distribution that can be non-convex, making averaging data dangerous as it can lead to task failure. Recent approaches address this by reformulating BC's policy as a generative process.

Denoising Diffusion Probabilistic Models (DDPM) \cite{ddpm} have emerged as a new class of generative models that outperform previous generative models. DDPM reverses a diffusion process that adds noise to a clean sample until it becomes Gaussian noise. By solving a Stochastic Differential Equation, it then generates a clean sample from this noise. Denoising Diffusion Implicit Models (DDIM) \cite{ddim} instead solve the reverse process as an Ordinary Differential Equation, reducing inference steps for faster computation at the expense of quality. Originally used for image generation, recent works \cite{janner2022planning, visuo_diffusion, 3d_diffusion,zhu2024diffusionmodelsreinforcementlearning} have applied these techniques to reinforcement and imitation learning. They generate action trajectories that mimic human demonstrations conditioned on the task's state, effectively capturing high-dimensional probability distributions and handling non-convex, non-connected distributions with multiple modes.

Flow Matching \cite{flow} is a novel generative method based on optimal transport theory \cite{albergo2023stochastic, albergo2023building}, sharing theoretical similarities with DDPM and DDIM. It is simpler with fewer hyperparameters and more numerically stable than DDIM. Flow Matching produces straighter paths in the transport flow, improving generation quality with a given number of integration steps, or enabling faster inference at equivalent quality using fewer steps, crucial for real-time robotic applications. In line with \cite{hu2023rfpolicy,hu2024adaflow,braun2024riemannian}, which demonstrated improvements of flow over diffusion in simulated robotic tasks, we investigate the application of Flow Matching in robotics and deploy it on real humanoid.

\begin{figure*}[t]
    \centering
	\includegraphics[trim=0cm 0cm 0cm 0cm,clip,width=0.95\linewidth]{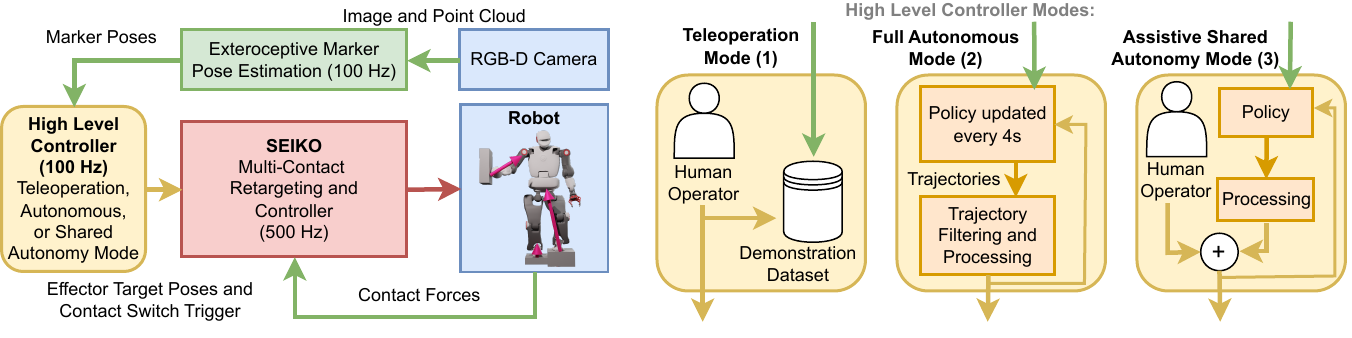}	
    \caption{
    The system architecture uses three different operational modes (right) as high-level controllers outputting effector commands. These commands are realized on the robot (left) by SEIKO Retargeting \cite{seiko,seiko2} and SEIKO Controller \cite{rouxel2024multicontact}.
    }
    \label{fig:architecture}
\end{figure*}

\section{Method}\label{sec:method}

We present a learning-from-demonstration approach for humanoid robots to perform multi-support manipulation tasks, enhancing manipulation capabilities through additional contacts and whole-body motions. These tasks are executed either autonomously or through assistive shared autonomy, where the human operator partially commands the robot while the learned policy provides assistance and contact placement. We highlight how our method handles contact switch transitions and controls the resulting multi-contact motions on real hardware.

\subsection{Overall Architecture}

We designed our architecture (Fig.~\ref{fig:architecture}) with two hierarchical modules to enhance robustness. A model-based low-level controller addresses whole-body optimization, multi-contact force distribution, contact switching and tracking with strict feasibility constraints. A learning-based high-level controller handles Cartesian effector commands, contact locations, and sequencing. It outputs a Cartesian pose target in world frame and a contact switch command for each effector. Effectors can either be fixed in contact with the environment (enabled state), actively applying forces to balance the robot, or not in contact and free to move (disabled state). The contact switch command is a discrete signal that triggers the transition between enabled and disabled states implemented by the low-level controller.

The high-level controller in Fig.~\ref{fig:architecture} operates in three different modes. The teleoperation mode is used to create a dataset recording effector commands sent to the low-level controller and poses of external markers detected by the robot's head camera. The human operator directly commands the robot to collect demonstrations, solving the task from randomized initial states or performing recovery actions from manually selected states outside nominal execution. The autonomous mode uses the policy trained by imitation of collected demonstrations to solve the task. The assistive shared autonomy mode combines human and policy commands to address out-of-distribution tasks. The operator commands one effector while the policy autonomously manages the others. The policy uses identical inputs and post-processing in both shared and full autonomous modes. However, in shared autonomy mode, the operator's commands replace the policy's output for the effector they control.

Despite \cite{visuo_diffusion,3d_diffusion} showed that diffusion-based imitation learning can learn from raw images or point clouds, we opt to use fiducial markers in this work to monitor the task's exteroceptive state. This allows us to focus instead on the challenges related to contact switches and multi-contact. An RGB-D camera on the robot's head detects these markers in the color image using the AprilTags system \cite{krogius2019flexible}. The 3D positions and orientations of the markers in the camera frame are extracted from the point cloud. These coordinates are then transformed into the robot's world frame using the forward kinematic model. The poses of the markers are recorded in the dataset during human expert demonstrations and fed as input to the autonomous policy.

\subsection{Behavioral Cloning Policy and Contact Switch}\label{sec:learning}

The behavioral cloning policy takes as input the current effector pose commands, contact states, and detected marker poses. It outputs a trajectory of future effector pose commands and contact switch commands for all effectors. Formally, the policy is defined as follows:
\begin{equation}\label{eq:policy}
\begin{aligned}
& \text{Policy } \pi: \bm{s}_k \longrightarrow \bm{a}_k \text{ where}\\
& \bm{s}_k \text{ includes: }
\begin{bmatrix}
    \bm{X}\up{eff i}_k &
    c\up{eff i}_k &
    \tau\up{eff i}_k
\end{bmatrix}~\forall i,~
\begin{bmatrix}
    \bm{X}\up{tag j}_k &
    \tau\up{tag j}_k
\end{bmatrix}~\forall j,\\
& \bm{a}_k \text{ includes: } 
\begin{bmatrix}
    \bm{X}\up{eff i}_k & \bm{X}\up{eff i}_{k+1} & \cdots & \bm{X}\up{eff i}_{k+N}\\
    \gamma\up{eff i}_k & \gamma\up{eff i}_{k+1} & \cdots & \gamma\up{eff i}_{k+N}\\
\end{bmatrix}~\forall i, \\
\end{aligned}
\end{equation}
and where
$i \in \N$ indexes the effectors,
$j \in \N$ indexes the markers,
$N \in \N$ is the number of predicted time steps,
$k \in \N$ is the inference time step,
$\bm{X}\up{eff i}_k \in \SE$ is the pose command in world frame of effector $i$ at time step $k$,
$c\up{eff i} \in \{0,1\}$ is the boolean contact state command of effector $i$ ($0$ for disabled or $1$ for enabled),
$\gamma\up{eff i}_k \in \R$ is the continuous contact state command (disabled or enabled) for effector $i$ at time step $k$,
$\tau\up{eff i} \in \R$ is the (clamped) elapsed time since last contact switch of effector $i$,
$\bm{X}\up{tag j} \in \SE$ is the latest updated pose estimate in world frame of marker $j$,
$\tau\up{tag j} \in \R$ is the (clamped) elapsed time since marker $j$ pose was last detected and its pose was updated.

\begin{figure}[t]
    \centering
	\includegraphics[trim=0cm 0cm 0cm 0cm,clip,width=\linewidth]{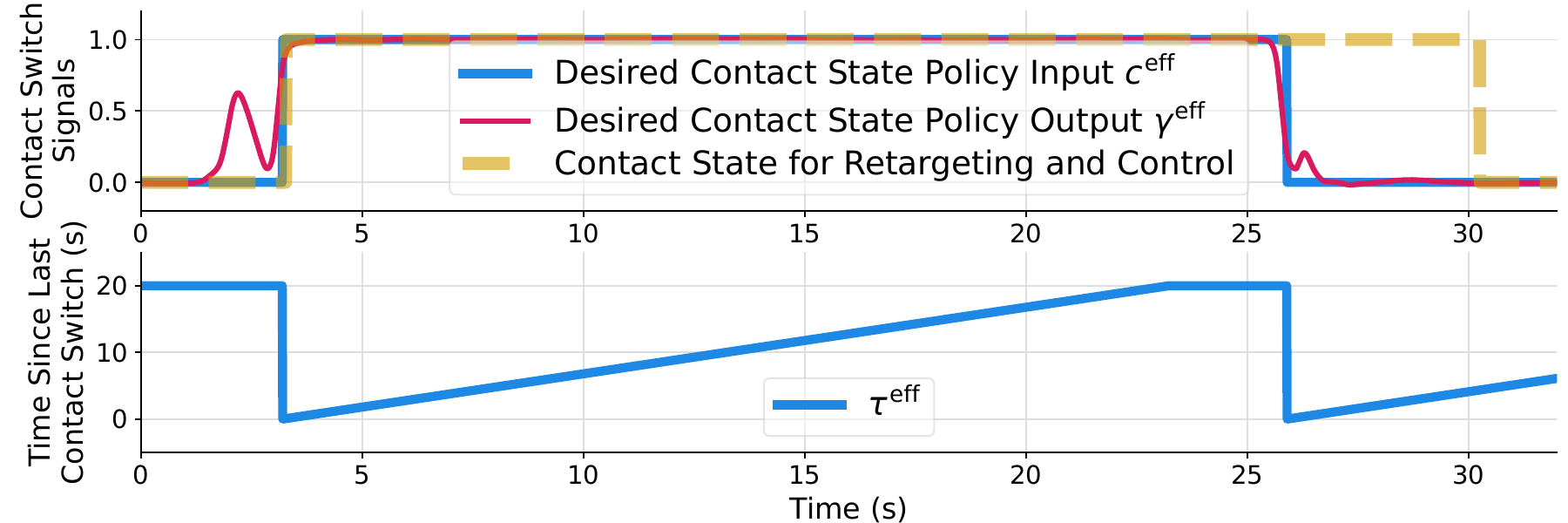}	
    \caption{Input and output signals used to command contact switches. The policy outputs the continuous command signal $\gamma\up{eff}$ converted to a discrete contact switch command $c\up{eff}$ using a hysteresis threshold. The time information $\tau\up{eff}$ disambiguates the states and produces the waiting behavior required when removing or adding a contact.
    }
    \label{fig:contact_switch}
\end{figure}

Fig.~\ref{fig:contact_switch} depicts the signals employed by the policy to implement contact switching commands. When adding or removing a contact, the low-level retargeting and controller necessitate a time delay to smoothly transfer the robot's weight and redistribute contact forces. The policy uses $\tau\up{eff i}$ to observe the progression of the contact transition and reproduce the waiting behaviors demonstrated by the operator upon triggering a contact switch. The policy outputs the continuous signal $\gamma\up{eff}$ indicating the desired state for a contact. The contact transition is activated and sent to the low-level whole-body retargeting module upon a state change of the discretized desired state $c\up{eff}$ defined by the following hysteresis threshold:
\begin{equation}
    c\up{eff i}_k = 
    \begin{cases}
    1 & \text{if } c\up{eff i}_{k-1} = 0 \text{ and } \gamma\up{eff i}_k \geqslant 0.8 \text{ and } \tau\up{eff i}_k \geqslant 20.0\\
    0 & \text{if } c\up{eff i}_{k-1} = 1 \text{ and } \gamma\up{eff i}_k \leqslant 0.2 \text{ and } \tau\up{eff i}_k \geqslant 20.0\\
    c\up{eff i}_{k-1} & \text{else}\\
    \end{cases}
\end{equation}
To avoid unbounded and out-of-distribution states, we clamp the time inputs  $\tau\up{eff}$ and $\tau\up{tag}$ to a maximum of $20$~s (see Fig.\ref{fig:contact_switch}) as a large upper bound. We apply data augmentation during training by randomizing detected marker times $\tau\up{tag}$ to enhance robustness against occlusions.

\begin{figure*}[t]
    \centering
    \includegraphics[trim=0cm 0cm 0cm 0cm,clip,width=0.95\linewidth]{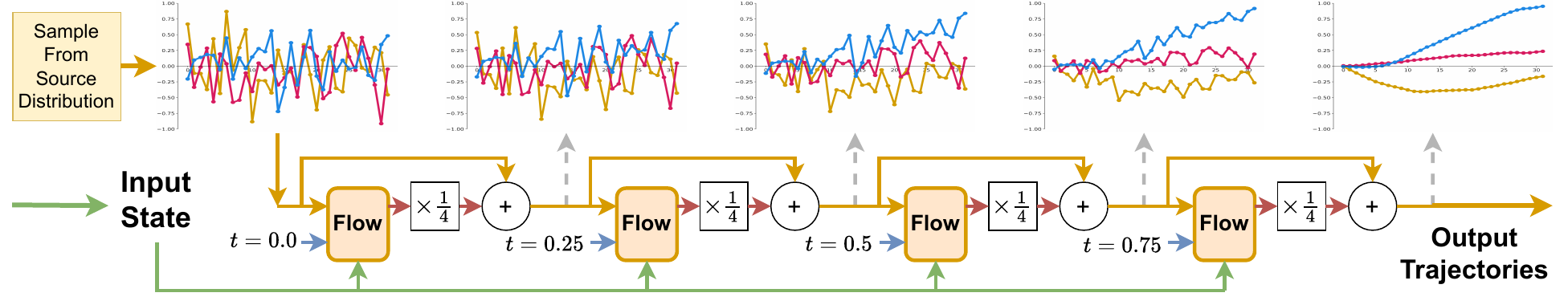}
    \caption{Inference of the policy through integration of the learned flow in 4 steps.
    }
    \label{fig:flow}
\end{figure*}

\subsection{Trajectory Generation with Flow Matching}\label{sec:flow}

We build the behavioral cloning policy as a generative process, which learns from data a probability distribution and sample new elements from it. The resulting policy is stochastic, and samples trajectories that mimic the ones demonstrated by the human operator in the same state. Specifically, we employ the Flow Matching method \cite{flow}, which constructs a \textit{flow} vector field that continuously transforms a \textit{source} probability distribution into a \textit{destination} distribution. Fig.~\ref{fig:concept} illustrates a flow transforming a 1D simple source distribution which can be easily sampled, into a more complex, multi-modal distribution. Flow Matching, grounded in optimal transport theory, can be seen as the deterministic counterpart to Diffusion methods \cite{albergo2023stochastic}. After sampling from the source distribution, the integration of the flow produces samples from the destination distribution deterministically, contrasting with Diffusion \cite{ddpm}, which introduces noise during transport. Flow Matching typically yields straighter flows, enabling faster inference.

The training of Flow Matching is defined as follows:
\begin{equation}
\begin{aligned}
    & \bm{a}\up{src} \sim \mathcal{P}\up{src}, ~\bm{a}\up{dst} \sim \mathcal{P}\up{dst}, ~t \sim \mathcal{U}[0,1], ~\bm{z}_t = (1-t)\bm{a}\up{src} + t\bm{a}\up{dst}, \\
    & \mathcal{L}\down{flow} = \mathbb{E}_{\bm{a}\up{src},\bm{a}\up{dst},t} \norm{f(\bm{z}_t, t, \bm{s}) - (\bm{a}\up{dst} - \bm{a}\up{src})}^2,
\end{aligned}
\end{equation}
where
$\mathcal{P}\up{src}$ is the source distribution, chosen as a multivariate normal distribution $\mathcal{P}\up{src} = \mathcal{N}(\bm{0}, \bm{I})$,
$\mathcal{P}\up{dst}$ is the destination distribution (here the demonstration trajectories),
$\bm{a}\up{src}$ and $\bm{a}\up{dst}$ are the trajectories sampled from source and destination distributions (see (\ref{eq:policy})),
$t \in \R$ is the scalar flow transport time uniformly sampled between $0$ and $1$ representing the progression of the transformation from source to destination,
$\bm{s}$ is the input state associated to the command trajectory $\bm{a}\up{dst}$,
$\bm{z}_t$ is the interpolated trajectory at transport time step $t$ between source and destination trajectories,
$\mathcal{L}\down{flow} \in \R$ is the scalar training loss function.
$f$ is the flow model conditioned by the state $\bm{s}$:
\begin{equation}
    \text{Flow } f: \bm{z},t,\bm{s} \longrightarrow \Delta \bm{z},
\end{equation}
implemented as a neural network and trained using back-propagation minimizing the loss $\mathcal{L}\down{flow}$.

The inference procedure is illustrated in Fig.~\ref{fig:flow}. A noisy trajectory is first sampled from the source distribution and then transformed into the destination trajectory by integrating the flow from $t=0$ to $t=1$ over several steps. Formally, the inference process is defined as follows:
\begin{equation}
\begin{aligned}
    & \bm{z}_0 = \bm{a}\up{src}, \bm{z}_1 = \bm{a}\up{dst} \\
    & \bm{z}_{t+\Delta t} = \bm{z}_t + \Delta t f(\bm{z}_t,t,\bm{s})
    & \text{for } t = 0 ... 1 \\
\end{aligned}
\end{equation}

\begin{figure}[t]
    \centering
	\includegraphics[trim=0cm 0cm 0cm 0cm,clip,width=\linewidth]{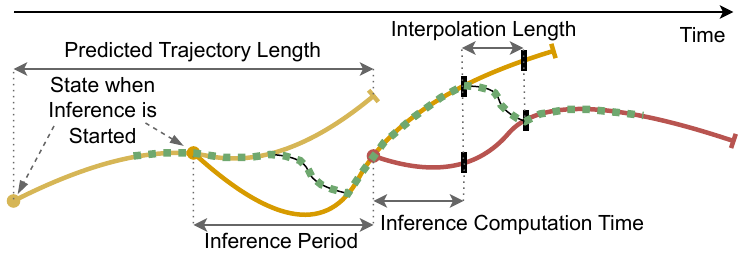}	
    \caption{The policy outputs new trajectories (yellow, orange, red) of fixed length at regular intervals. Due to the time required for inference, each new trajectory is seamlessly stitched online with the previous one to prevent discontinuities using linear interpolation. The dashed green trajectory depicts the resulting commands sent to the low-level retargeting.}
    \label{fig:trajectories}
\end{figure}

\subsection{Trajectories Stitching and Processing}

Since inference is not instantaneous and the policy outputs a trajectory of future commands, online stitching (Fig.~\ref{fig:trajectories}) and processing are required to ensure smoothness, safety, and robustness of the commands sent to the low-level controller. The autonomous high-level controller computes the policy's next commands trajectory in parallel while continuously sending effector commands to the SEIKO low-level, sampled from the previous trajectory. When a new inference starts, the policy uses the latest effector commands and marker pose estimates as the current state. Upon completion, a smooth transition to the new commands trajectory is achieved through linear interpolation over a fixed time. The policy produces trajectories represented as $5$~Hz time series, which are resampled at $100$~Hz using linear interpolation for use by the high-level controller. A zero-phase low-pass filter (first-order exponential filter) is applied on the trajectory to remove residual noise from flow inference and interpolation.

\subsection{Multi-Contact SEIKO Retargeting and Controller}\label{sec:seiko}

Robots with multiple limbs in multi-contact exhibit redundancy both in kinematics and contact force distribution. Many different whole-body postures can achieve a desired effector pose, and many contact force distributions can maintain equilibrium for a given posture. To perform multi-support manipulation on real robots, it is essential to consider kinematic and actuator torque limits, contact and balance constraints to prevent slipping, failing, and ensure operational safety. Contact switch transitions are discrete decisions that significantly impact system's balance, requiring careful consideration for smoothness and safety. These transitions are not always feasible and typically take a few seconds. Smoothly removing a contact requires gradually reducing the contact force to zero by adjusting the whole-body posture and redistributing the contact forces, necessitating precise regulation of the contact forces on the actual system. Our proposed method relies on the SEIKO (Sequential Equilibrium Inverse Kinematic Optimization) Retargeting and Controller methods developed in our previous work \cite{seiko,seiko2,seiko3,rouxel2024multicontact} to address these diverse challenges.

SEIKO Retargeting \cite{seiko,seiko2} uses a model-based Sequential Quadratic Programming (SQP) optimization to compute a feasible whole-body configuration (joint positions and contact forces) tracking the effector pose commands. It integrates the command filtering pipeline detailed in \cite{seiko2,rouxel2024multicontact}. The retargeting adapts the robot's motion to enforce safety constraints in response to risky or infeasible commands from either human operator or the policy. 

SEIKO Controller \cite{rouxel2024multicontact} integrates an explicit modeling of joint flexibility and utilizes an SQP whole-body admittance formulation to regulate the contact forces on a position-controlled humanoid robot. The controller improves robustness to model errors and enable real robot experiments by regulating contact forces. To further enhance robustness against inaccuracy in contact placement, we extended SEIKO Controller with the effector admittance control scheme named ``damping control'' detailed in \cite{murooka2022centroidal}. This scheme addresses scenarios where the learned policy activates a contact too early while still in the air, or too late after already exerting forces on the environment. The presentation, comparison, and discussion of this control scheme are provided in the supplementary material of \cite{rouxel2024multicontact}.

\section{Results}\label{sec:results}

\begin{table}[t]
\centering
\caption{Training, inference and processing hyperparameters}
\label{table:parameters}
\begin{tabular}{|ll|} 
    \hline
    Description & Value\\
    \hline
    Trajectory sub-sampling frequency & $5$~Hz\\
    Trajectory length & $N=32$ steps ($6.4$~s)\\
    Input state dimension & $20+10\times\text{\#markers}$\\
    Output trajectory dimension & $13$\\
    Elapsed times $\tau\up{eff}$ and $\tau\up{tag}$ clamping & $20$~s\\
    U-net model layer sizes & $[32, 64, 64]$\\
    U-net model convolution kernel size & $5$\\
    Number of model's trainable parameters & $666029$ (single marker)\\
    Training learning rate & $10^{-4}$\\
    Number of training epochs & $5000$\\
    Inference period & $4$~s\\
    Interpolation length & $0.5$~s\\  
    \hline
\end{tabular}
\end{table}

\subsection{Implementation Details}\label{sec:implementation}

The Talos robot is a humanoid robot manufactured by PAL Robotics of $1.75$~m height, $99.7$~kg and $32$ degrees of freedom. We replaced the robot's right-hand gripper and forearm with a 3D-printed, ball-shaped hand to withstand high force contact. In our experiments, we control only $22$ joints, all in position-control mode, excluding those in the neck, forearms, and wrists. We mounted an Orbbec Femto Bolt RGB-D camera on the robot's head, replacing the original camera and providing color images and point clouds.

In our experiments, the robot's left and right hands are used as effectors commanded by the operator for manipulation tasks, while the feet remain fixed. Only the right hand is used for making contact with the environment on the 3d-printed ball shape. Depending on the experiment, we use between 1 and 3 external fiducial markers. The human operator teleoperates the robot with a direct line of sight, and uses separate 6-DoF input devices\footnote{3Dconnexion SpaceMouse: {\url{https://3dconnexion.com/uk/spacemouse/}}} to command the velocity of each hand, providing after integration the effector pose commands.

The policy is trained in Python using the PyTorch library with GPU acceleration, whereas online inference is performed in C++ on the CPU (Intel i9-9880H $2.30$~GHz). See Table~\ref{table:parameters} for hyperparameters. The flow model is implemented as a 1D convolutional U-Net neural network with residual connections, akin to the model implemented\footnote{Diffusion Policy code: \url{https://github.com/real-stanford/diffusion_policy}} in \cite{visuo_diffusion}. For each effector, the predicted poses in the output trajectories $(\bm{X}\up{eff i}_l)_{l=k}^{k+N}$ are encoded relative to the pose in the input state $\bm{X}\up{eff i}_k$ such that all predicted positions and orientations trajectories start from zero. The effector orientations in the input state are encoded using the 6D rotation representation \cite{zhou2019continuity}, whereas the relative orientations in the predicted output trajectories are expressed as 3D axis-angle vectors.

SEIKO Retargeting and Controller are implemented\footnote{SEIKO implementation: \url{https://github.com/hucebot/seiko_controller_code}} in C++, using the Pinocchio rigid body library, the QP solver QuadProg \cite{goldfarb1983numerically} and run onboard the robot at $500$~Hz. The fiducial external markers are detected in the color image using the AprilTag library at $30$~Hz.

\begin{figure*}[t]
    \centering
    \begin{minipage}[c]{5.5cm}
        \includegraphics[trim=0.5cm 2cm 2.5cm 1.2cm,clip,width=0.4\linewidth]{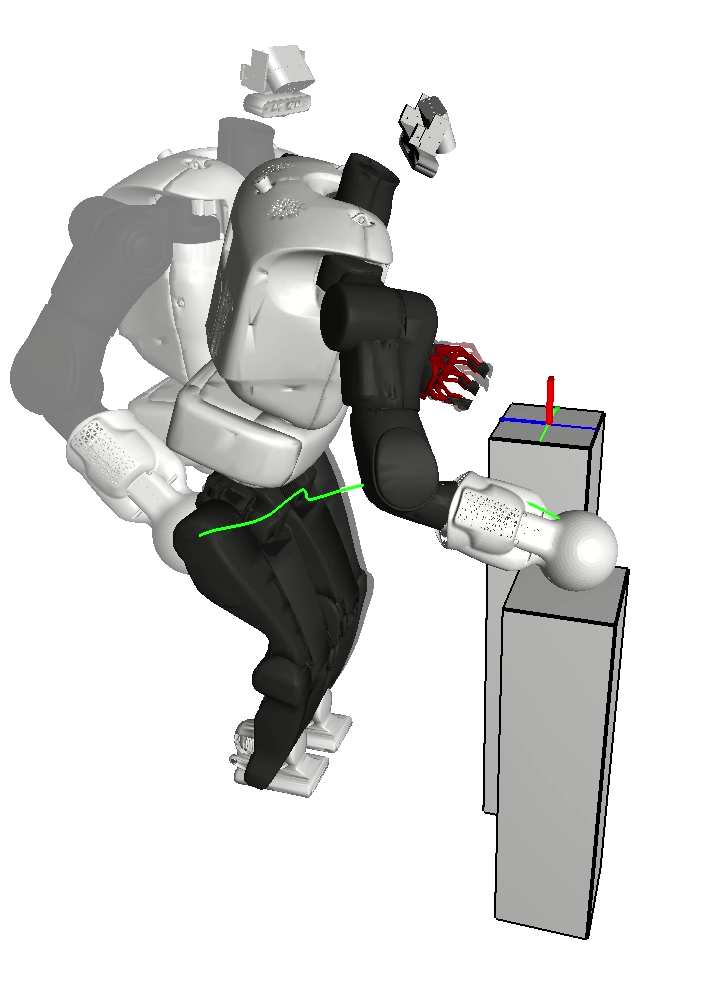}%
        \includegraphics[trim=2.0cm 2cm 1.0cm 1.2cm,clip,width=0.4\linewidth]{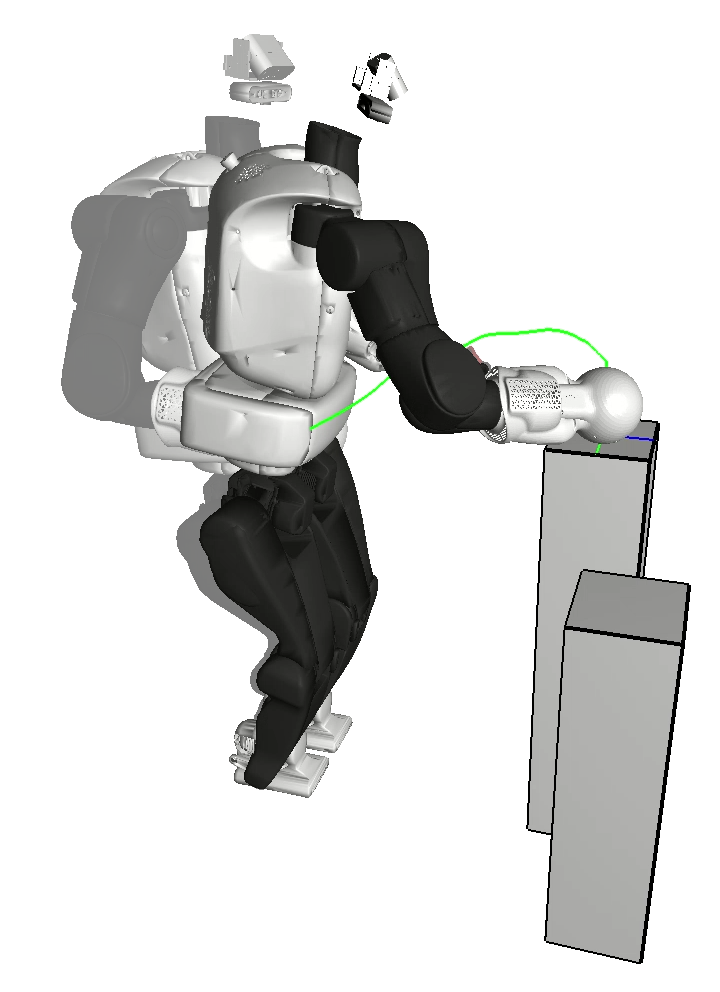}%
    \end{minipage}
    \centering
    \small
    \begin{tabular}{|c|c|cc|cc|}
        \hline
        \multirow{2}{*}{Method} & \multirow{2}{*}{\thead{Inference\\Time\\(ms)}} & \multicolumn{2}{c|}{In Distribution} & \multicolumn{2}{c|}{Out of Distribution}\\
        \cline{3-6}
        & & \thead{Success\\Rate} & \thead{Median [Q1, Q3]\\(cm)} & \thead{Success\\Rate} & \thead{Median [Q1, Q3]\\(cm)}\\
        \hline
        Demonstrations & -- & $100\%$ & $1.3$ [$0.9$, $2.0$] & -- & --\\
        \hline
        Flow $20$ steps & $35\pm4$ & $99\%$ & $1.4$ [$1.0$, $2.1$] & $\bm{78\%}$ & $\bm{3.4}$ [$\bm{1.9}$, $\bm{5.7}$]\\
        \hline
        DDPM $100$ steps & $178\pm12$ & $100\%$ & $1.5$ [$0.9$, $1.8$] & $69\%$ & $4.0$ [$2.4$, $5.9$]\\
        \hline
        DDIM $20$ steps & $39\pm4$ & $100\%$ & $1.4$ [$0.9$, $2.0$] & $67\%$ & $3.9$ [$2.7$, $6.1$]\\
        \hline
        Supervised Learning & $\bm{3\pm1}$ & $92\%$ & $4.1$ [$2.6$, $5.3$] & $52\%$ & $7.6$ [$4.3$,  $12.3$]\\
        \hline
    \end{tabular}
    \caption{Simulated contact reaching task: The robot extends its right hand to establish contact with one of the two support platforms (left), where the initial position of the hand and the position of the platforms relative to the robot are randomized. Autonomous mode results over 100 trials for each model, both in and out of distribution, are reported (right). Success rate (contact switch activated and distance to closest platform $<8$~cm), median, Q1, and Q3 quartiles of contact placement error (if contact was established) are detailed.
    }
    \label{fig:task_reach}
\end{figure*}

\subsection{Simulated Reaching and Contact Placement Task}

 We compare in our experiments the Flow Matching method for robotics applications with its Diffusion counterpart and a classical supervised learning baseline. We present statistical results for the following variant methods:
\begin{itemize}
    \item Demonstrations: dataset collected by the expert human operator and used to train all autonomous policies.
    \item Flow $20$ steps: Flow Matching method described in Section~\ref{sec:flow}. The flow is integrated (see Fig.~\Ref{fig:flow}) over $20$ steps.
    \item DDPM $100$ steps: vanilla DDPM method \cite{ddpm} trained with $100$ denoising steps, and inferred with $100$ steps.
    \item DDIM $20$ steps: uses the same trained model as DDPM, but is inferred with the Diffusion implicit variant \cite{ddim} and $20$ steps, expected to be faster than DDPM at the expense of quality.
    \item Supervised Learning: classical behavior cloning method \cite{argall2009survey} trained with Mean Square Error (MSE) loss. It has same inputs-outputs and also predicts trajectories, but it is not a generative process and does not capture the data distribution.
\end{itemize}

We evaluate the main capabilities of policies: first, their ability to autonomously perform contact switching as described in Section~\ref{sec:learning}; second, their ability to learn from demonstrations with a multi-modal distribution; third, their accuracy in placing contacts, which is crucial for robotics applications; and fourth their inference time. Within the simulated task illustrated in Fig.~\ref{fig:task_reach}, we teleoperated $86$ demonstrations totaling $2442$~s. The hand was placed randomly on either the left or right platform, regardless of the initial state to create a bimodal distribution. An external marker is attached on top of the left platform, with the position of the right platform fixed relative to the left. We also assess how the policies generalize out-of-distribution, where initial hand positions and platform positions are uniformly sampled from a wider range that encompass and excludes the range used for in-distribution cases.

Both Flow and Diffusion approaches outperform the baseline (Fig.~\ref{fig:task_reach}), as supervised behavior cloning is hindered by the multi-modal nature of the distribution, causing the baseline to average out across non-convex spaces. Flow Matching also slightly outperforms Diffusion in out-of-distribution cases, with favorable inference time and accuracy, which is in line with other works published on this topic \cite{hu2023rfpolicy, hu2024adaflow, braun2024riemannian}.

\begin{figure*}[t]
    \centering
	\includegraphics[trim=0cm 0cm 0cm 0cm,clip,width=0.16\linewidth]{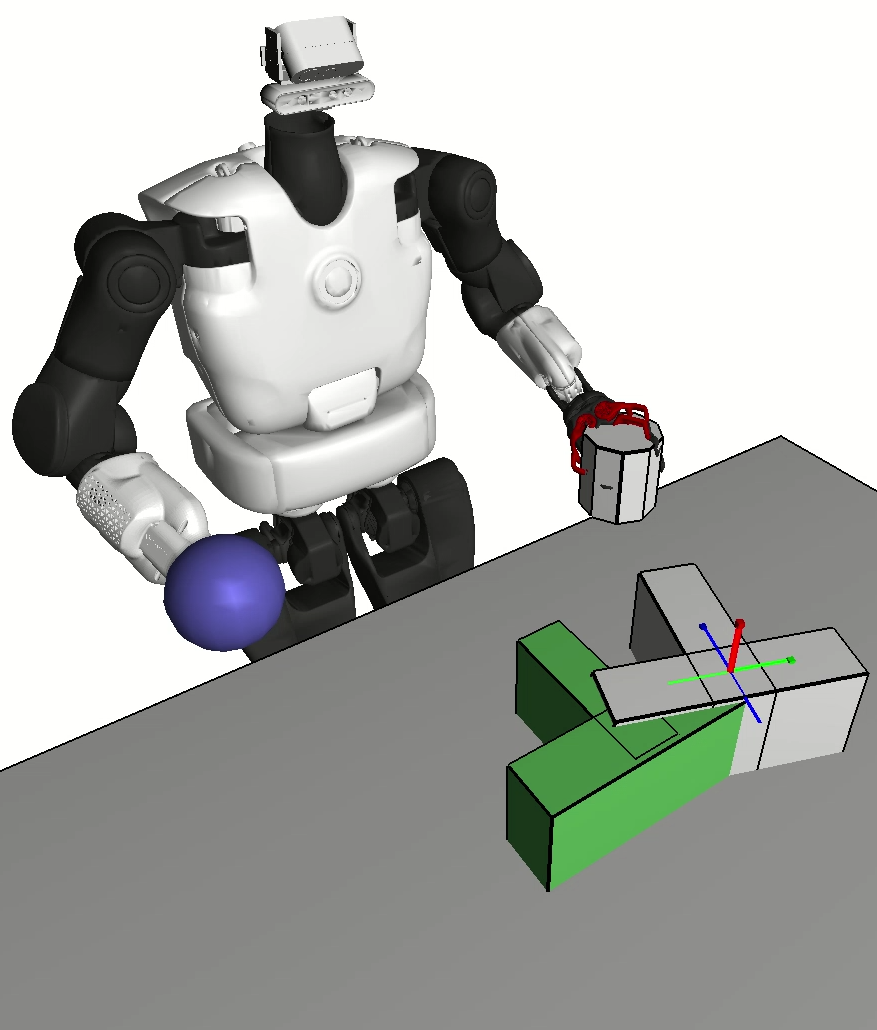}
    \includegraphics[trim=0cm 0cm 0cm 0cm,clip,width=0.16\linewidth]{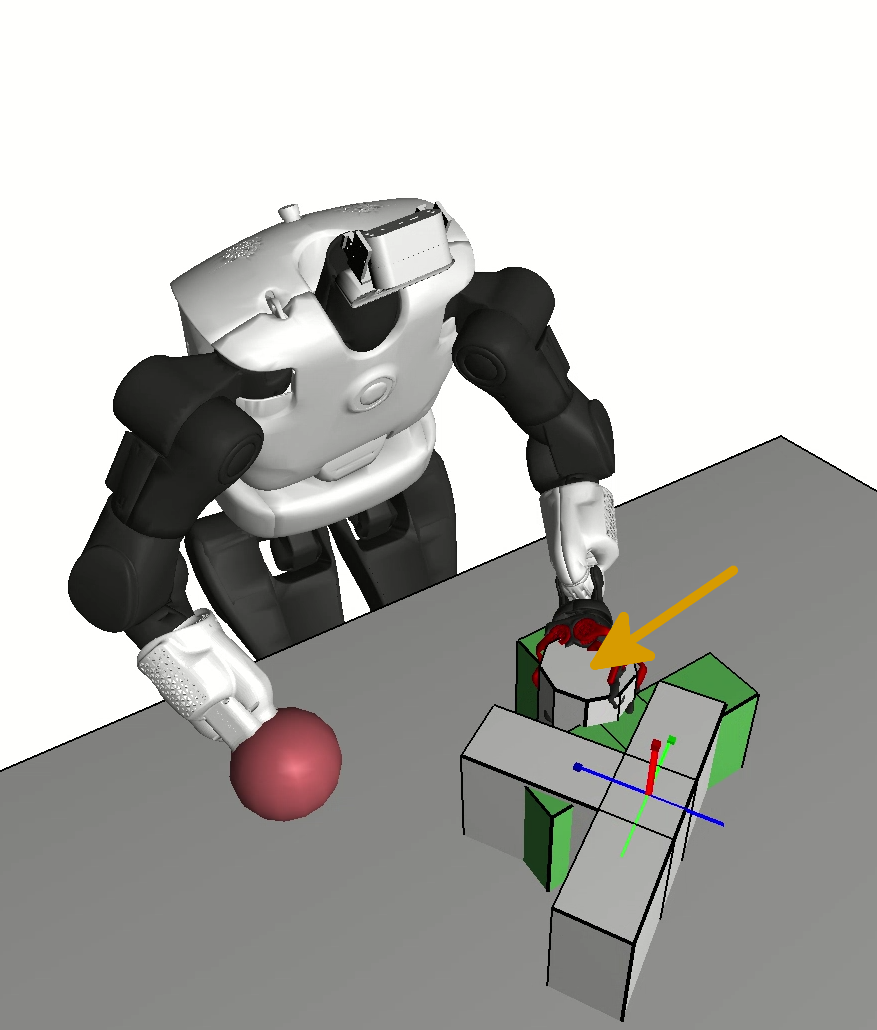}
    \includegraphics[trim=0cm 0cm 0cm 0cm,clip,width=0.16\linewidth]{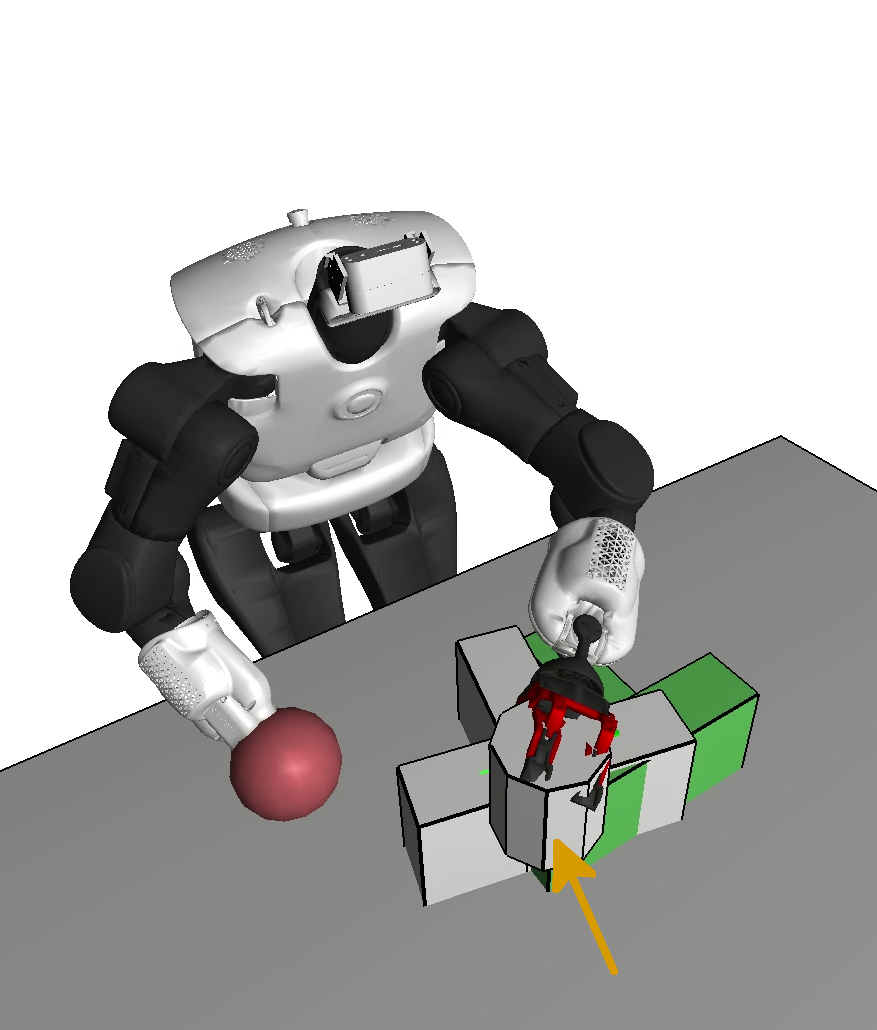}
    \includegraphics[trim=0cm 0cm 0cm 0cm,clip,width=0.16\linewidth]{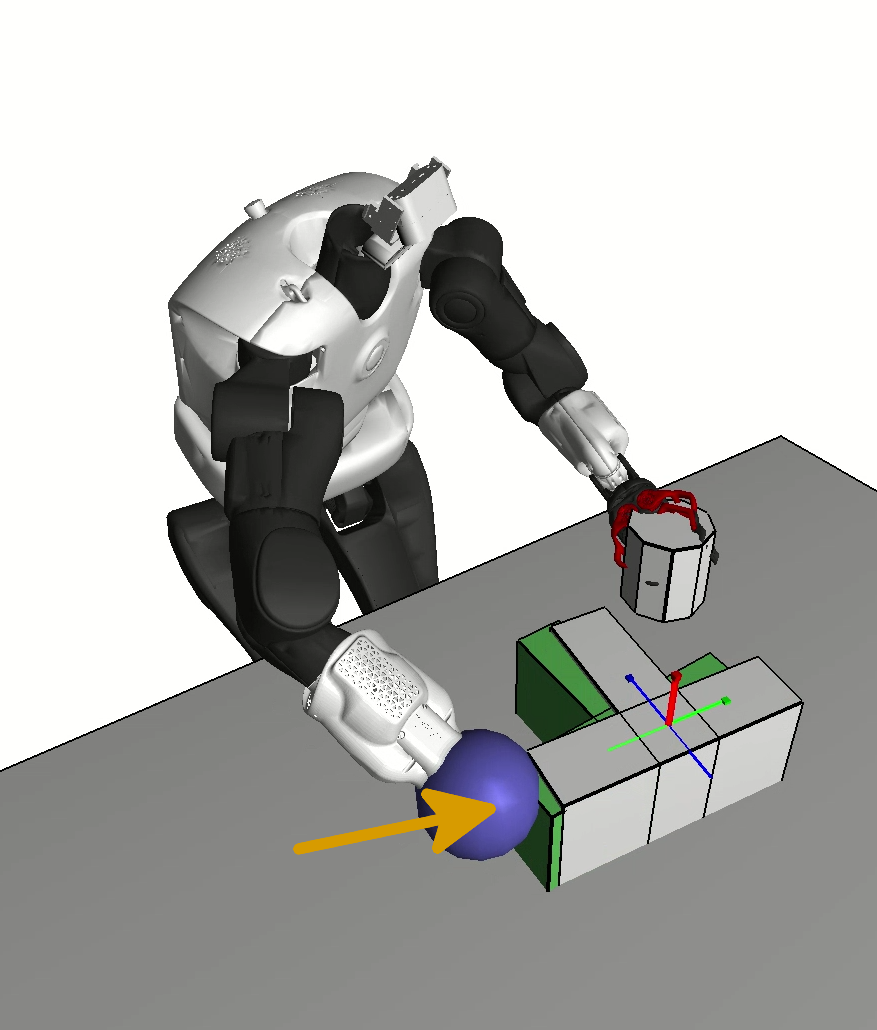}
    \includegraphics[trim=0cm 0cm 0cm 0cm,clip,width=0.16\linewidth]{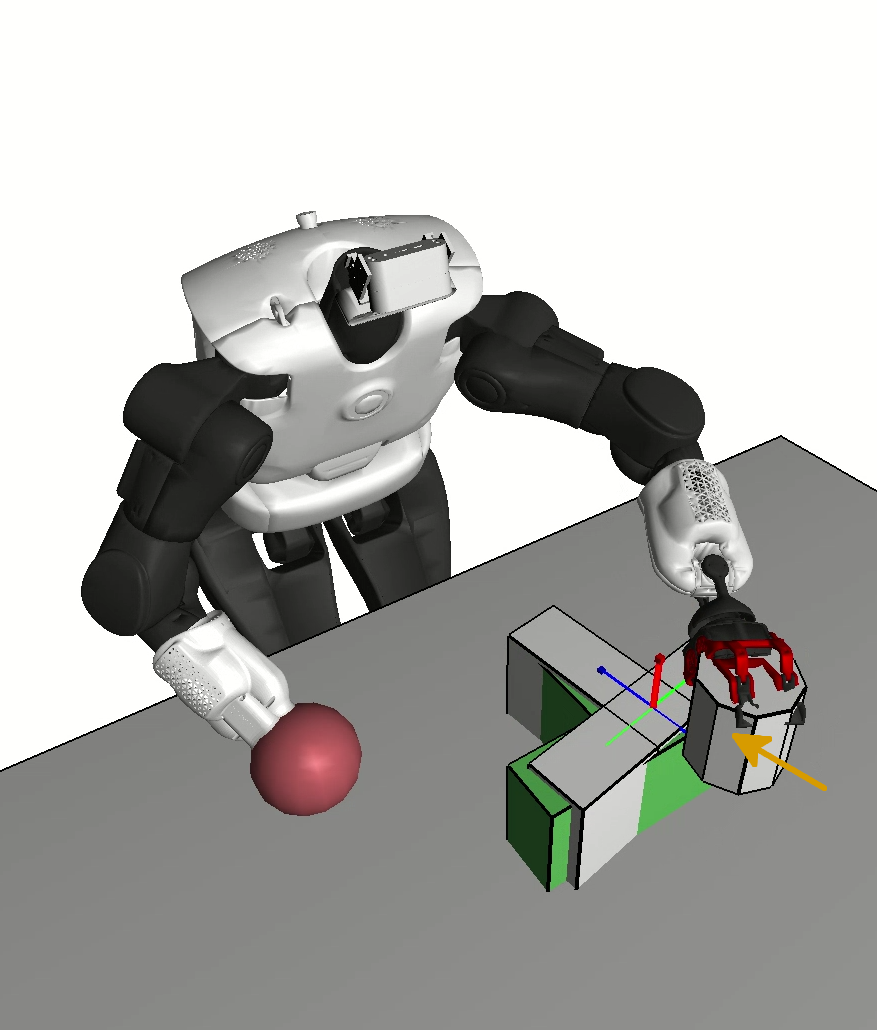}
    \includegraphics[trim=0cm 0cm 0cm 0cm,clip,width=0.16\linewidth]{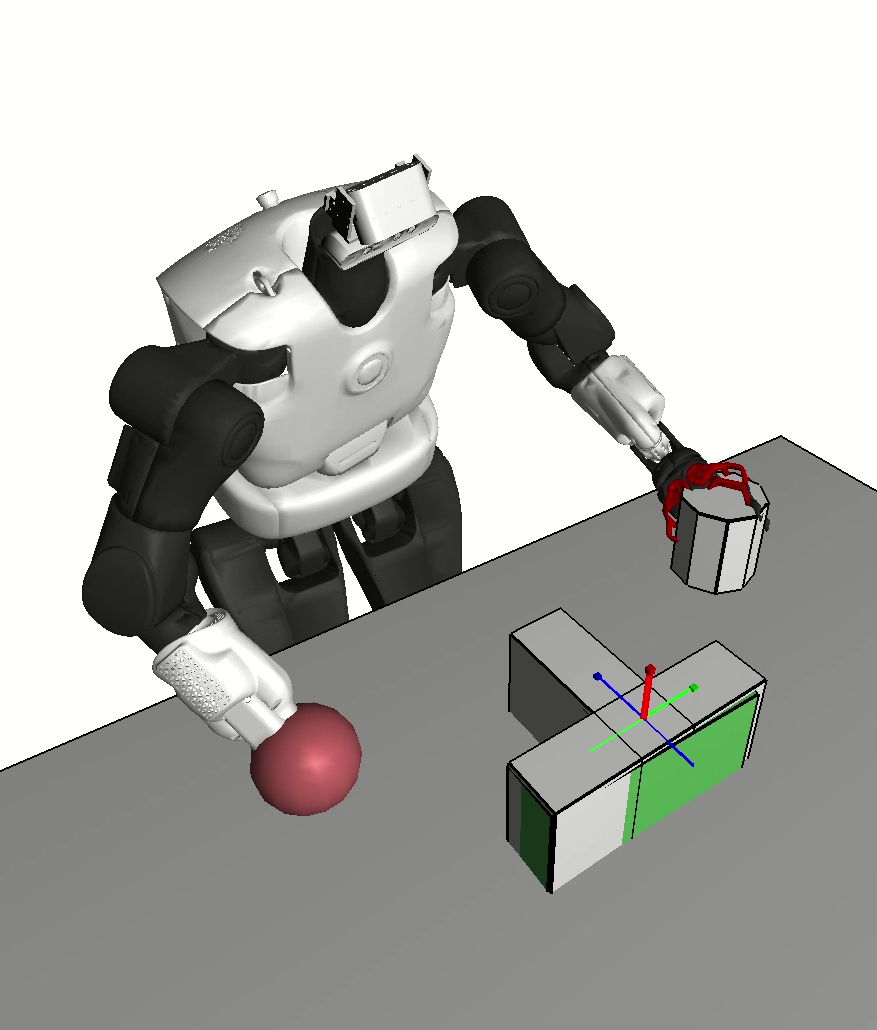}
    \caption{Simulated non-prehensile pushing task: the robot uses both hands to push the gray T-shaped object to the match the green target shape. For further pushes from behind the object, the robot extends its reach by placing its right hand on the table for support (red color indicates that the contact is enabled, blue is disabled).
    }
    \label{fig:task_push}
\end{figure*}

\begin{figure}[t]
    \centering
	\includegraphics[trim=0cm 0cm 0cm 0cm,clip,width=\linewidth]{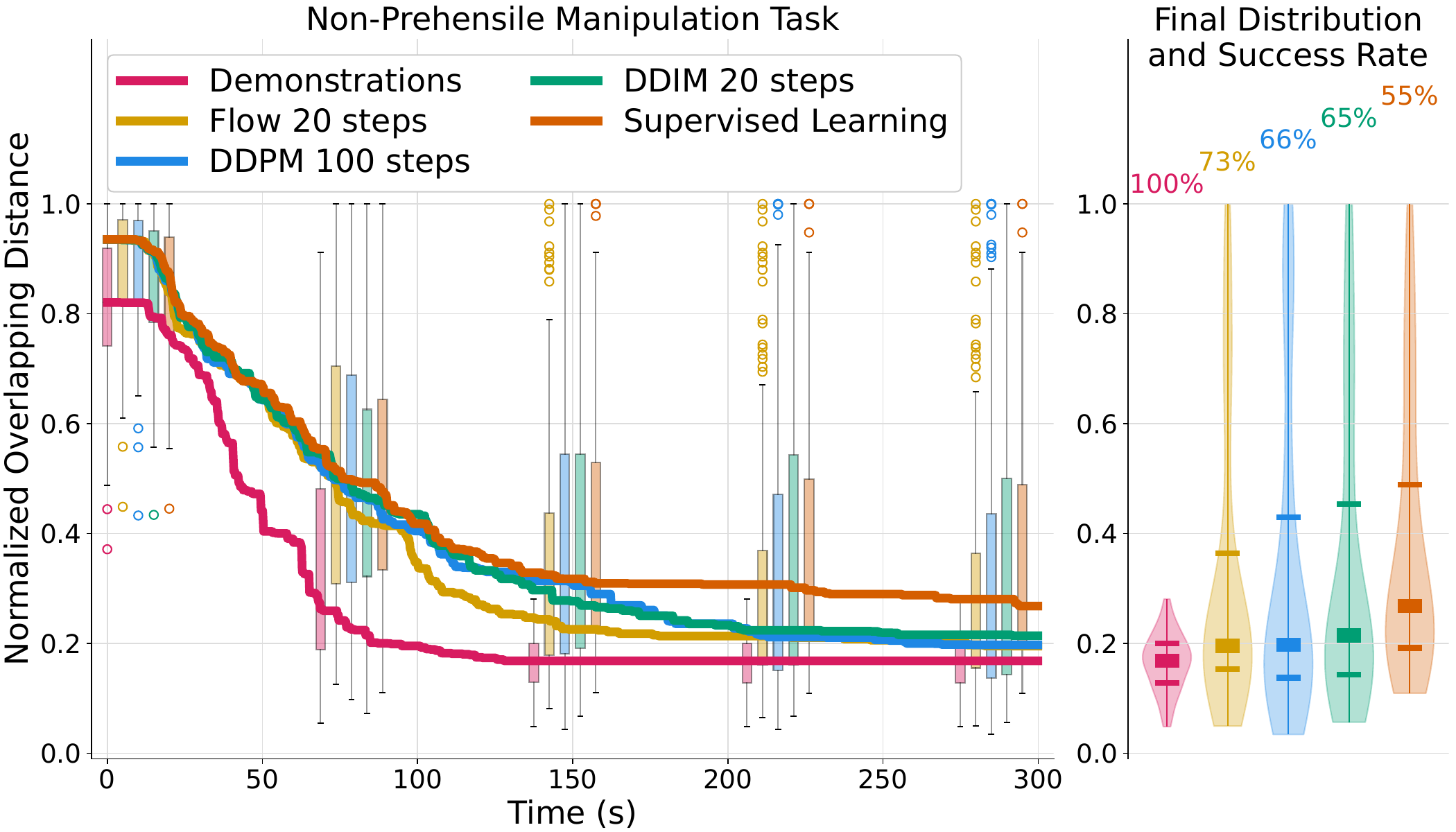}	
    \caption{
        Simulated T-box multi-support pushing task:  normalized overlapping distance between the object and the target pose over time (left) and final success rate (distance < $0.3$) after $300$~s (right). The distribution, median, and quartile statistics, evaluated over 100 trials for each method, are reported.
    }
    \label{fig:stats_push}
\end{figure}

\subsection{Simulated Non-Prehensile Manipulation Task}

We then evaluated our proposed method on the more challenging non-prehensile manipulation task shown in Fig.~\ref{fig:task_push}. This task aims to thoroughly test multi-support and whole-body strategies with higher multi-modality, necessitating both the addition and removal of contacts. The humanoid robot must use both hands to push a concave T-shaped 3D object on a planar table surface, maneuvering it to match a target position and orientation fixed on the table. Solving the task strongly relies on multi-support capabilities, as the robot can not reach forward far enough to push the object from behind without using its right hand as additional support. The robot interacts with the box using contact-rich dynamics that heavily depend on geometries of the box and robot's effector, as well as friction and sliding properties of surfaces. The task allows for various multi-modal strategies by applying different pushing sequences on the box's sides. It requires several contact switches to push the box left and right with both hands, followed by precise final adjustments. This box-pushing task is a more challenging 3D whole-body extension of a simpler 2D top-down environment used as a benchmark in previous work \cite{visuo_diffusion, moura2022non, ferrandis2023nonprehensile}.

In a real-time simulated environment, we teleoperated the robot to record $68$ demonstrations totaling $6161$~s. The initial position and orientation of the box were randomized on the table, while both the target pose for the object and the position of the robot's feet remained fixed. A single marker is placed and attached on top of the object, providing its pose to the policy. After training, the resulting policies were evaluated in the simulated environment for $300$~s and across $100$ trials each.

We quantify the task performance of how the box's pose matches the target by measuring the planar overlapping surface between the manipulated T-shaped object and the fixed T-shaped target of the same size. Specifically, we define the task error metric as the normalized overlapping distance $\sqrt{1-(\text{overlapping surface})/(\text{shape surface})}$, where an error of 0.0 indicates a perfect match, and an error of 1.0 indicates no overlap between the two shapes. Since policies lack stopping criteria and continuously interact with the object, we consider the lowest error achieved so far within each trial.

Fig.~\ref{fig:task_push} showcases an example of autonomous execution, while Fig.~\ref{fig:stats_push} presents the comparison statistical results. All autonomous methods compared have failure cases. The two most common failure scenarios are when the robot collides with the top of the box, considered as a stopping criterion, or mistakenly pushes the box into a configuration where necessary adjustments are no longer reachable. The Flow and Diffusion methods, theoretically very similar, exhibit comparable behavior and performance by the end of episodes. Both methods outperform the supervised behavioral cloning baseline. Flow method tends to marginally outperform its Diffusion counterparts which is coherent with \cite{hu2024adaflow,hu2023rfpolicy}, achieving statistically faster task completion and exhibiting slightly less variance. As expected, Diffusion with 100 steps performs marginally better than Diffusion 20 steps, showcasing the trade-off between inference time and performances.

\begin{figure*}[t]
    \centering
	\includegraphics[trim=0cm 0cm 0cm 0cm,clip,width=0.19\linewidth]{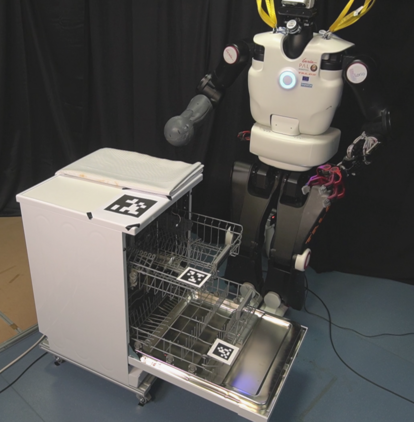}
 	\includegraphics[trim=0cm 0cm 0cm 0cm,clip,width=0.19\linewidth]{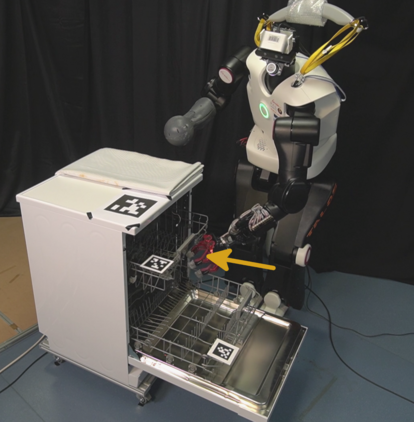}
  	\includegraphics[trim=0cm 0cm 0cm 0cm,clip,width=0.19\linewidth]{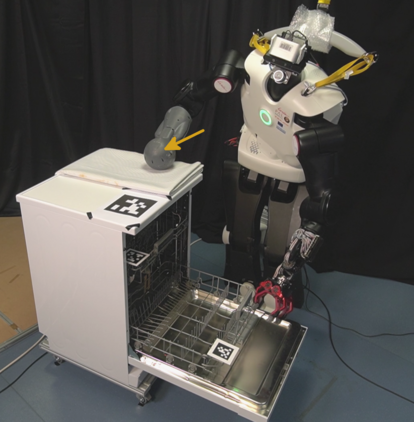}
   	\includegraphics[trim=0cm 0cm 0cm 0cm,clip,width=0.19\linewidth]{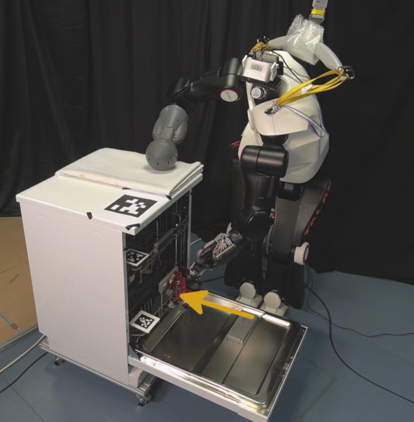}
    \includegraphics[trim=0cm 0cm 0cm 0cm,clip,width=0.19\linewidth]{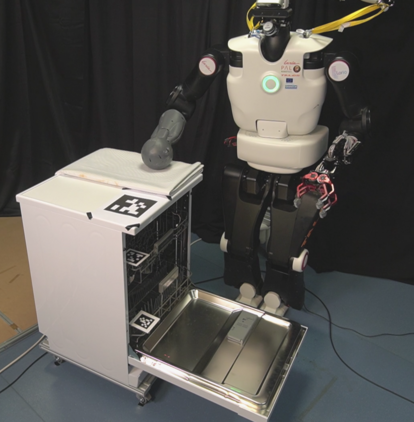}
    \caption{
    Talos robot closes the upper and lower drawers of a dishwasher with its left hand. To maintain balance while bending forward to close the lower drawer, the robot places its right hand on top of the dishwasher. The task is performed in both fully autonomous mode and shared autonomy mode. In shared autonomy mode, the human operator commands the left hand while the robot automatically places the right hand in contact when the operator attempts to reach the lower drawer.
    }
    \label{fig:task_dishwasher}
\end{figure*}

\begin{figure*}[t]
    \centering
	\includegraphics[trim=0cm 0cm 0cm 0cm,clip,width=0.33\linewidth]{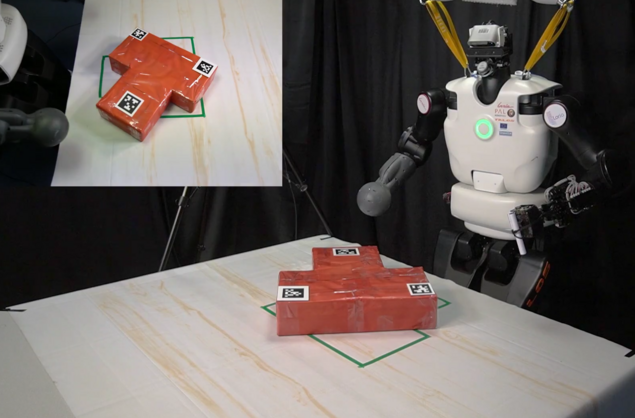}%
 	\includegraphics[trim=0cm 0cm 0cm 0cm,clip,width=0.33\linewidth]{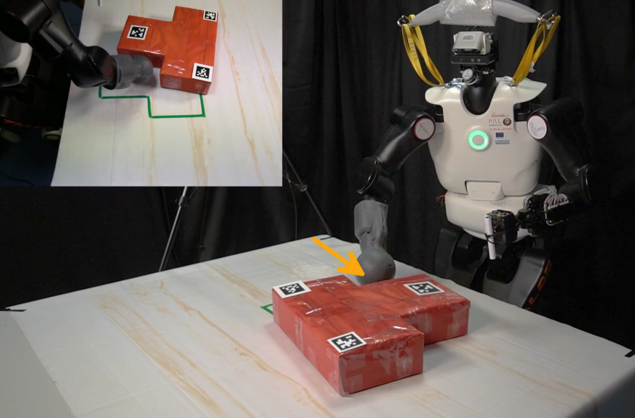}%
  	\includegraphics[trim=0cm 0cm 0cm 0cm,clip,width=0.33\linewidth]{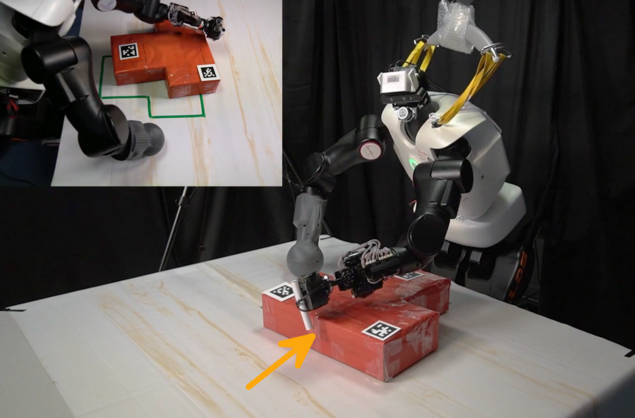}
    \includegraphics[trim=0cm 0cm 0cm 0cm,clip,width=0.33\linewidth]{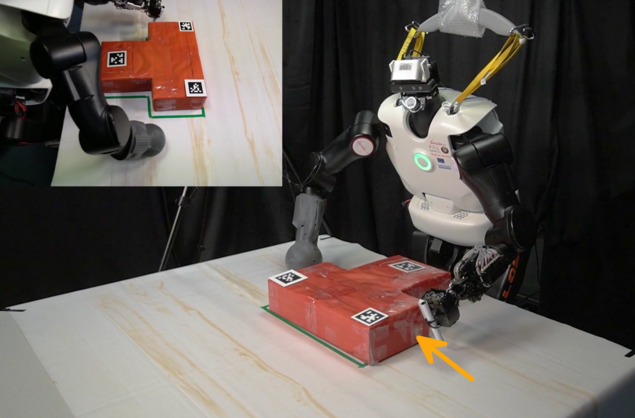}%
   	\includegraphics[trim=0cm 0cm 0cm 0cm,clip,width=0.33\linewidth]{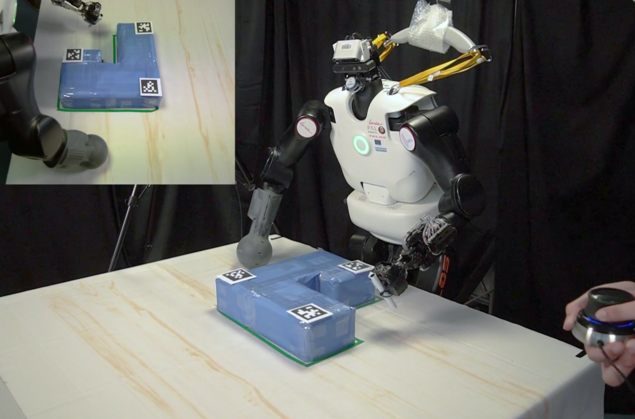}%
    \includegraphics[trim=0cm 0cm 0cm 0cm,clip,width=0.33\linewidth]{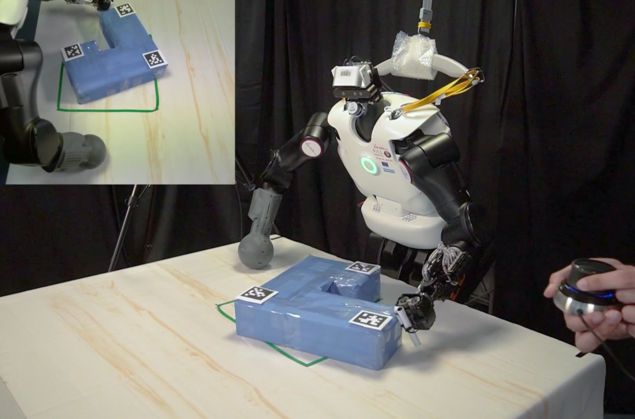}
    \caption{
    Non-prehensile box pushing task on the Talos robot. The Flow Matching policy, learned from demonstrations using the red T-box, can autonomously maneuver the red box to match the target. However, when applied out-of-distribution to the blue U-box, the autonomous mode alone cannot solve the task. Despite this, the policy still provides valuable assistance in shared autonomy mode, where the operator commands the left hand and the policy commands the right hand.
    }
    \label{fig:task_realpush}
\end{figure*}

\subsection{Hardware Experiments}

The attached and additional videos\footnote{Additional videos: \url{https://hucebot.github.io/flow_multisupport_website/}} showcase our multi-support hardware experiments on the Talos humanoid robot.

\subsubsection{Autonomous Mode In Distribution}

First, we validated our proposed contact placement and switching capabilities by having the robot push and close both upper and lower drawers of a dishwasher (Fig.~\ref{fig:concept}%
, Fig.~\ref{fig:task_dishwasher}%
). This task is straightforward for humans but remains challenging for humanoid robots. Because of both leg and torso joint position and torque limits, the robot needs to add an extra contact on top of the dishwasher to reach without falling the lower drawer, which is $40$~cm above the ground.

We teleoperated the robot to collect $35$ demonstrations with a total length of $1734$~s. Three markers were used, one on top of the dishwasher and one on each drawer. As shown in the additional videos, the reactive policy learned with Flow Matching successfully solves the task autonomously and responds to disturbances that may reopen already closed drawers. The robot first closes the upper drawer in double support and then reaches to close the lower drawer, placing additional right hand support on top of the dishwasher. This experiment also validates our architectural choice, where the low-level retargeting and controller successfully execute the multi-support manipulation tasks commanded by the learned policy. Without the controller enabled, any far-reaching motion tends to cause the robot to fall due to model errors.

Second, we demonstrated the box pushing task (Fig.~\ref{fig:task_push}) on the real Talos robot (Fig.~\ref{fig:concept}%
, Fig.~\ref{fig:task_realpush}%
). 
The robot easily manipulates the box with both hands when nearby. But, when pushing it from behind, the right hand contact is needed to compensate for interaction forces from the box's mass and static friction on the nearly fully extended left arm. We recorded $51$ demonstrations of $5521$~s with the red T-shaped box, and we used three markers on the object instead of only one to mitigate sensor noise and self-occlusion. The policy learned with Flow Matching successfully solves the red T-shape case using both hands, dynamically adding or removing right-hand contacts, and effectively reacts to disturbances applied to the object (see additional videos). 

\subsubsection{Assistive Shared Autonomy Out of Distribution}

Imitation learning only performs well in-distribution for the task it was trained on. We assessed this by testing the box-pushing task with a blue U-shaped box, representing an out-of-distribution case. As expected, the autonomous policy trained on red T-shape performed poorly with the blue U-shaped box, failing and getting stuck while attempting to push on non-existent sides.

We evaluated our assistive shared autonomy mode \cite{li2023classification} (Fig.~\ref{fig:architecture}) on the real robot (see additional videos) aiming to address this known downside of imitation learning. In this mode, the human operator commands only the left hand, while the policy commands the right hand, which is responsible for adding or removing the upper body support. We solved the box pushing task in the blue U-shape out-of-distribution case using this assisted teleoperation approach. The operator makes fine adjustments with the left hand while the policy adds the right-hand contact to enable distant reach. When the object moves right and becomes unreachable with the left hand, the policy removes the right-hand contact, and pushes the object back toward the left side. In the dishwasher task, the shared autonomy mode automatically places the right-hand contact on top of the dishwasher when the operator commands the left hand to go below a certain height while attempting to reach down.

\section{Discussion and Conclusion}

Our experiments with multi-support tasks show that Diffusion and Flow Matching both outperform the traditional behavior cloning with supervised learning approach (our baseline). We hypothesize that this advantage arises because these tasks are diverse, multi-modal, and require intricate strategies. The learned policies are robust enough to be deployed on a real, full-size humanoid robot (Talos), enabling it to autonomously perform multi-support tasks, including pushing a box and closing drawers with the help of the free hand for balance. Using a shared-autonomy assisted teleoperation approach, extendable to natural language interaction \cite{totsila2024words2contactidentifyingsupportcontacts}, we demonstrated that policies learned from demonstrations can assist in automatic contact placement, even for tasks that differ from the demonstrations.

Like other methods based on behavior cloning, the performances depends on the quality of expert demonstrations. The human operator must not only demonstrate the desired behavior but also include recovery actions that enable the policy to correct deviations from the nominal path and handle potential disturbances. Recent prior work \cite{visuo_diffusion,3d_diffusion} have shown that autonomous policies can be learned directly from raw images or point clouds, eliminating the need for fiducial markers. This aligns with our work and represents a natural extension, complementing the contact switch capability we propose. Our SEIKO low-level controller supports creating contacts using both predefined hand and foot effectors, while the imitation learning framework enables interaction with the robot’s entire geometry, such as non-prehensile pushing with its forearms. This paves the way for complex multi-support loco-manipulation tasks and further exploration of learning contact placement from human expertise.


\bibliographystyle{IEEEtran}
\bibliography{references}

\end{document}